# Multi-Agent Framework for Audit Risk Assessment with Explicit Uncertainty and Evidence Conflict Modeling

Yuhan Wang
Columbia University
New York, USA

Manqing Wang
Trine University
Phoenix, USA

Yixuan Lu
University of Sofia
Palo Alto, USA

Zhaoyue Peng
Westcliff University
San Francisco, USA

Shengda Lin*
University of Illinois at Urbana-Champaign
Champaign, USA

***Abstract*-Audit risk assessment increasingly benefits from combining heterogeneous evidence sources, yet existing approaches typically produce point predictions without quantifying how well different evidence streams agree. We propose UMAR (Uncertainty-Aware Multi-Agent Risk Assessment), a framework that employs three specialized agents-an MD&A Text Agent, a Financial Ratio Agent, and a CAM Agent-each producing independent risk scores with calibrated uncertainty estimates. An Uncertainty Aggregator based on Dempster–Shafer evidence theory fuses these scores while explicitly measuring inter-agent conflict. We evaluate UMAR on a U.S. dataset of 3,200 firm-year observations from SEC 10-K filings (2019–2023), with financial restatement as the target label. Experimental results show that UMAR achieves an AUROC of 0.782 and a PR-AUC of 0.341, outperforming logistic regression, XGBoost, FinBERT, and single- and dual-agent LLM baselines. UMAR attains the lowest expected calibration error (ECE of 0.052) among all methods and identifies evidence-conflict patterns that correlate with actual restatement risk, offering auditors potentially actionable and interpretable risk signals.**

## I. INTRODUCTION

The detection of material misstatements in financial reports is a central objective of the external audit process. Traditional audit risk assessment relies on auditors' professional judgment to integrate multiple evidence sources, but the growing volume and complexity of information motivate the exploration of computational approaches.

Machine learning methods have been applied to fraud detection and audit opinion prediction with considerable success. Bao et al. [1] demonstrated that ensemble learning models using raw accounting numbers outperform logistic regression for fraud prediction. Text-based approaches leveraging financial-domain NLP models such as FinBERT [2] have improved sentiment extraction from corporate filings. More recently, Lu et al. [3] combined MD&A text analysis with financial ratios in an LLM-based dual-agent framework for audit opinion prediction, finding that multi-source combinations outperform single-source models.

Despite these advances, several gaps remain. First, the EDINET-Bench evaluation [4] revealed that even advanced LLMs only marginally outperform logistic regression on complex financial tasks such as fraud detection, suggesting that naive application of general-purpose LLMs remains insufficient. Second, field studies have identified transparency, bias, robustness, and auditor over-reliance as critical barriers to AI adoption in auditing [5], [6]. Third, existing systems focus on classification accuracy but rarely quantify *uncertainty* or identify *evidence conflicts* across sources-information that is essential for auditors to make informed decisions.

We address these gaps by proposing UMAR (Uncertainty-Aware Multi-Agent Risk Assessment), a framework with three contributions: (1) a multi-agent architecture where each agent specializes in a distinct evidence source (MD&A text, financial ratios, and critical audit matters); (2) an uncertainty aggregation mechanism based on Dempster–Shafer evidence theory [7] that detects and flags inter-agent conflicts; and (3) an experimental evaluation on U.S. SEC 10-K filings demonstrating that explicit uncertainty modeling improves both predictive performance and calibration compared to existing approaches.

## II. METHODOLOGY FOUNDATION

Growing interest in evidence-centric intelligence systems has encouraged the development of architectures that preserve the unique informational characteristics of heterogeneous data sources throughout the reasoning process. Linguistic evidence embedded within financial and audit disclosures has increasingly been recognized as a structured source of risk-relevant information that can be modeled independently from numerical indicators. Artificial-intelligence-based analyses of Key Audit Matters demonstrate that audit-report narratives contain rich informational signals capable of supporting risk-oriented inference, while subsequent investigations into the determinants of linguistic characteristics further reveal the stability and interpretability of these textual representations across reporting environments [8], [9]. Complementing this perspective, knowledge-guided generative frameworks for financial fraud detection illustrate the value of maintaining explicit semantic structures during reasoning, and generative distribution modeling further emphasizes the importance of representing risk through probabilistic beliefs rather than deterministic outputs [10], [11]. Collectively, these methodological developments establish the foundation for treating heterogeneous audit evidence as distinct yet complementary sources of information and motivate the preservation of source-specific confidence information throughout the assessment process.

Functional decomposition has emerged as an influential design principle in intelligent decision-support systems, enabling specialized analytical modules to capture different dimensions of complex prediction tasks. Research on causality-aware memory architectures demonstrates that structured reasoning benefits from maintaining interpretable intermediate representations capable of tracing the contribution of individual evidence components to final decisions [12]. Similar principles are reflected in retrieval systems that preserve and reuse evidence across multiple reasoning stages, thereby improving consistency and transparency in multi-step inference [13]. Advances in temporal-structural representation learning further show that distributed information sources can be modeled through coordinated analytical units while retaining their local semantic characteristics [14]. Resource-efficient adaptation strategies for large language models provide additional support for modular architectures in which specialized components are optimized for distinct information domains without sacrificing overall system effectiveness [15]. Together, these studies provide the methodological basis for constructing a multi-agent framework in which independent agents analyze MD&A narratives, financial indicators, and audit disclosures as specialized evidence streams.

Capturing semantic, structural, and contextual dependencies across distributed observations remains a central objective of contemporary representation learning research. Multiscale temporal modeling demonstrates that predictive signals frequently emerge from interactions among observations operating at different resolutions and contextual levels [16]. Contrastive dependency learning extends this view by showing that explicit modeling of relationships among heterogeneous entities improves anomaly identification and representation quality [17]. Structure-aware graph-enhanced learning similarly highlights the value of preserving relational information during feature construction and inference [18]. Related investigations of neural architectures combining frequency-domain and attention-based mechanisms reveal how complementary representations can be integrated to capture diverse behavioral patterns within complex systems [19]. These methodological streams collectively reinforce the importance of maintaining modality-specific representations prior to aggregation, thereby supporting an architecture in which heterogeneous evidence is processed independently before higher-level fusion.

Uncertainty-aware inference has evolved from a supplementary evaluation criterion into a central component of modern risk modeling frameworks. Adaptive federated learning research demonstrates that heterogeneous analytical environments benefit from aggregation mechanisms capable of preserving local uncertainty while combining complementary knowledge across distributed models [20]. Similar observations arise from federated fine-tuning approaches for large language models, where semantic alignment across domains enables coordinated reasoning while retaining source-specific expertise [21]. Counterfactual multimodal risk modeling further illustrates the value of integrating heterogeneous evidence streams while explicitly accounting for uncertainty associated with each source of information [22]. These developments collectively support the formulation of uncertainty as a first-class modeling construct and provide a methodological rationale for adopting evidence-theoretic fusion mechanisms that jointly evaluate confidence, agreement, and conflict among independently generated assessments.

Interpretability, calibration, and causal consistency increasingly serve as key requirements for intelligent decision-support systems operating in high-stakes environments. Research on causal inference and exposure-bias correction demonstrates that explicitly modeling underlying causal structures contributes to more reliable predictive behavior and more meaningful decision support [23]. Consistency-aware causal learning further reinforces the importance of preserving coherent relationships among evidence sources during optimization and inference [24]. Context-aware ranking frameworks provide examples of balancing multiple information signals through structured integration procedures rather than relying on isolated predictors [25]. Structural prompting techniques reveal that intermediate reasoning structures can substantially improve transparency and reasoning quality within language-model-based systems [26]. Complementary advances in large-scale multi-model coordination illustrate how specialized computational components can operate collaboratively toward a unified objective while preserving their individual strengths [27]. Taken together, these methodological directions establish the conceptual and technical foundation of UMAR, supporting a framework that combines specialized evidence analysis, uncertainty-aware prediction, and structured evidence fusion to generate calibrated and interpretable audit risk assessments.

# III. METHODOLOGY

## A. Framework Overview

UMAR consists of four components operating on a firm-year observation $(x_i, y_i)$, where $x_i$ comprises the MD&A text $T_i$, financial statement data $F_i$, and CAM disclosures $K_i$, while $y_i \in \{0, 1\}$ indicates whether a financial restatement occurred within two years. Fig. 1 illustrates the overall architecture.

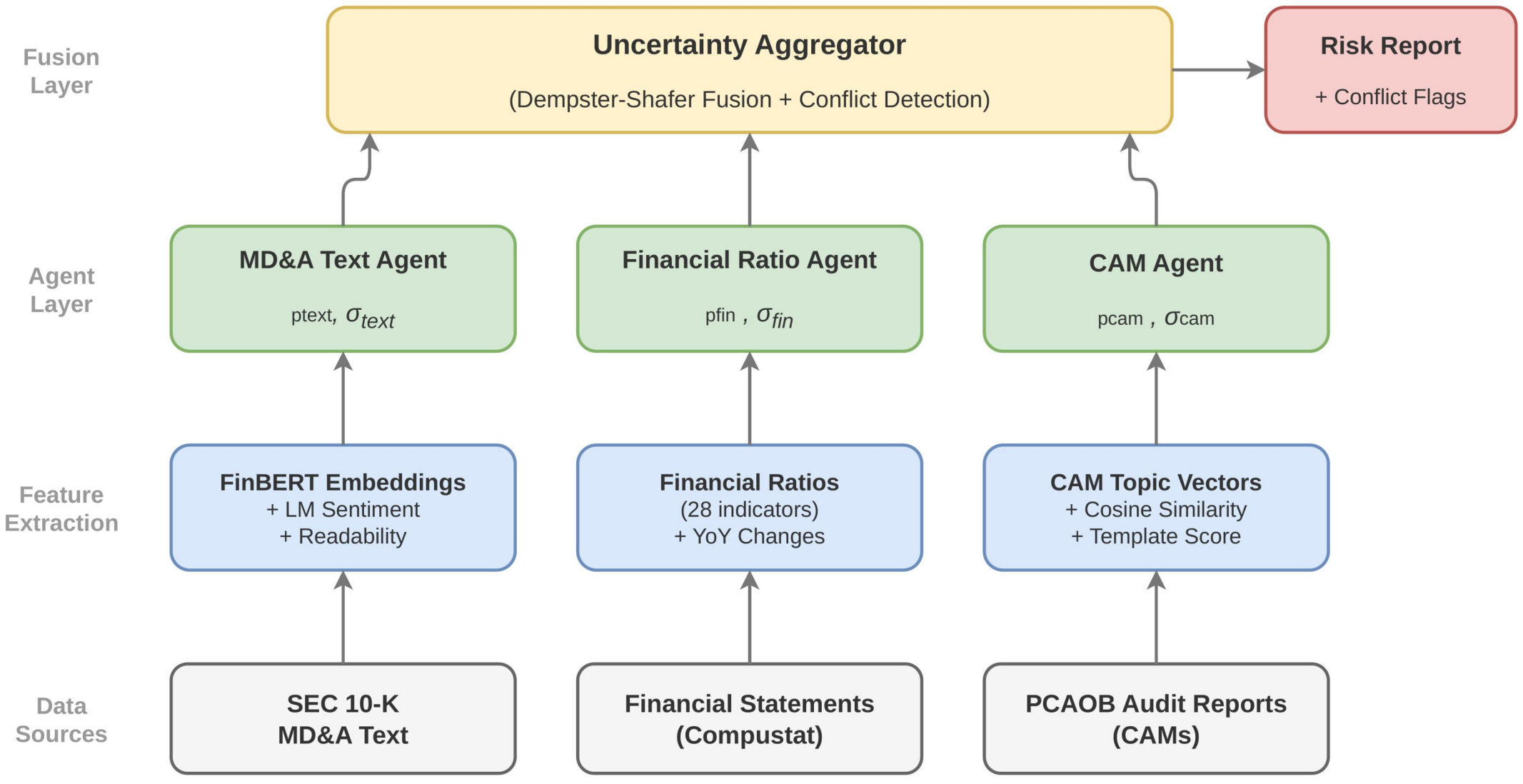

Fig. 1. Architecture of the UMAR framework. Three specialized agents extract calibrated risk probabilities and uncertainty estimates from heterogeneous data sources. The Uncertainty Aggregator fuses agent outputs via Dempster–Shafer theory and flags evidence conflict patterns.

### B. MD&A Text Agent

This agent processes the MD&A section from 10-K filings. We extract three feature groups: (i) FinBERT embeddings capturing contextual semantics, (ii) Loughran–McDonald sentiment ratios (negative, uncertainty, litigious word proportions), and (iii) readability metrics (Fog Index, document length). These features are fed into a gradient-boosted classifier that outputs a calibrated risk probability $p_{\text{text}} \in [0,1]$ via Platt scaling on a held-out calibration set.

### C. Financial Ratio Agent

This agent ingests 28 financial indicators from Compustat, organized into five categories: profitability (ROA, ROE, operating margin), leverage (debt-to-equity, interest coverage), liquidity (current ratio, quick ratio), cash flow quality (operating cash flow to net income), and anomaly indicators (accounts receivable days change, inventory turnover change, accruals ratio). An XGBoost classifier produces a calibrated risk probability $p_{\text{fin}}$ via isotonic regression calibration.

### D. CAM Agent

This agent analyzes Critical Audit Matters from PCAOB audit reports. We extract: (i) topic vectors via FinBERT embeddings of each CAM paragraph, (ii) year-over-year cosine similarity $s_{\text{cos}}$ between consecutive reports to detect template-like disclosures, and (iii) a divergence score measuring whether CAM topics have changed while financial conditions deteriorated. Because CAM-derived features are relatively sparse, the agent applies Platt scaling on the calibration set to output $p_{\text{cam}}$.

### E. Uncertainty Estimation and Aggregation

Each agent's calibrated probability $p_j$ is used to compute an uncertainty score $\sigma_j$ defined as the normalized prediction entropy:

$$\sigma_j = \frac{-p_j \log p_j - (1-p_j)\log(1-p_j)}{\log 2} \#(1)$$

where $\sigma_j \in [0,1]$, with $\sigma_j = 1$ indicating maximum entropy ($p_j = 0.5$) and $\sigma_j = 0$ indicating minimum entropy ($p_j \in \{0,1\}$). Probabilities are clipped to $[\epsilon, 1-\epsilon]$ with $\epsilon = 10^{-6}$ to avoid numerical issues.

Given the three agent outputs $(p_j, \sigma_j)$ for $j \in$ {text,fin,cam}, the aggregator constructs basic probability assignments (BPAs) within the Dempster–Shafer framework. For each agent $j$, we define:

$$m_j(H) = p_j \cdot (1-\sigma_j), \quad m_j(L) = (1-p_j)\cdot(1-\sigma_j) \#(2)$$

$$m_j(\Theta) = \sigma_j \#(3)$$

where $H$ denotes high risk, $L$ denotes low risk, and $\Theta = \{H, L\}$ represents the frame of discernment. The mass $m_j(\Theta)$ captures the agent's residual uncertainty.

The combined belief is obtained via Dempster's rule of combination:

$$m_{1,2}(A) = \frac{1}{1-\kappa} \sum_{B\cap C=A} m_1(B)\cdot m_2(C) \#(4)$$

where $\kappa = \sum_{B\cap C=\emptyset} m_1(B)\cdot m_2(C)$ quantifies the conflict between two evidence sources. The three agents are combined sequentially, and the final conflict indicator is $\kappa^* = \max\{\kappa_{\text{text,fin}}, \kappa_{\text{text,cam}}, \kappa_{\text{fin,cam}}\}$. When $\kappa^* > \tau$, the observation is flagged for human review ($\tau = 0.3$, selected on the calibration set). The final fused risk score is the pignistic probability:

$$\text{BetP}(H) = m(H) + \frac{1}{2} m(\Theta) \#(5)$$

where $m(\{H\})$ and $m(\Theta)$ are the fused masses for high risk and the full frame, respectively.

## IV. Experiments

### A. Dataset and Setup

We construct a dataset from U.S. large accelerated filers during 2019–2023. We restrict to fiscal years ending on or after

June 30, 2019, because PCAOB AS 3101 required CAM disclosures for large accelerated filers starting from this date. Table I summarizes the sample construction process.

TABLE I.
DATASET CONSTRUCTION PIPELINE.

| Step | Firm-years |
|---|---|
| SEC 10-K filings (FY end ≥ Jun 2019) | 6,842 |
| Matched with Compustat financial data | 5,517 |
| Matched with PCAOB audit reports (CAMs) | 4,283 |
| Matched with Audit Analytics restatements | 3,891 |
| Random subsample (stratified by year) | 3,200 |

MD&A text is extracted from 10-K filings via SEC EDGAR, financial data from Compustat, and CAM disclosures from PCAOB audit reports. The restatement label is derived from the Audit Analytics Non-Reliance Restatement database, resulting in an 8.2% positive rate (263 restatement firm-years). We use a chronological split: training on 2019–2021 (1,920 observations), calibrating on 2022 (640), and testing on 2023 (640). The two-year outcome window for the 2023 test set has fully elapsed by 2025, ensuring label completeness.

## B. Baselines

We compare UMAR against seven baselines: (1) Logistic Regression using all features; (2) Random Forest (500 trees, max depth 8); (3) XGBoost (300 rounds, learning rate 0.05, max depth 6); (4) FinBERT fine-tuned on MD&A text (3 epochs, batch size 16, learning rate $2 \times 10^{-5}$, max 512 tokens); (5) Longformer (max 4,096 tokens, 3 epochs); (6) Single-LLM using GPT-4o with concatenated inputs (temperature 0, max 1,024 output tokens); (7) Dual-Agent LLM following Lu et al.

## C. Main Results

Table II presents the main results. UMAR achieves the highest AUROC (0.782) and PR-AUC (0.341), outperforming the strongest baseline (Dual-Agent LLM) by 4.1 and 4.6 percentage points, respectively, with the lowest Brier Score (0.068) and ECE (0.052). Bootstrap resampling (1,000 iterations) confirmed UMAR ranking first in 94.2% of resamples. The Single-LLM baseline performs below XGBoost, consistent with findings from EDINET-Bench that general-purpose LLMs do not reliably outperform traditional methods on structured financial tasks without domain-specific adaptation.

TABLE II.
COMPARISON OF RESTATEMENT PREDICTION PERFORMANCE ON THE 2023 TEST SET. BEST RESULTS ARE BOLD; SECOND-BEST ARE UNDERLINED.

| Method | AUROC | PR-AUC | Brier ↓ | ECE ↓ |
|---|---|---|---|---|
| Logistic Regression | 0.672 | 0.214 | 0.089 | 0.112 |
| Random Forest | 0.713 | 0.259 | 0.081 | 0.094 |
| XGBoost | 0.738 | 0.287 | 0.076 | 0.083 |
| FinBERT | 0.721 | 0.271 | 0.079 | 0.089 |
| Longformer | 0.709 | 0.262 | 0.082 | 0.095 |
| Single-LLM | 0.695 | 0.237 | 0.085 | 0.108 |
| Dual-Agent LLM | <u>0.741</u> | <u>0.295</u> | <u>0.074</u> | <u>0.079</u> |
| UMAR (Ours) | **0.782** | **0.341** | **0.068** | **0.052** |

## D. Top-k Risk Recall Analysis

Fig. 2 shows the recall of restatement firms at various inspection thresholds. At $k = 10\%$, UMAR identifies 50% of all restatement firms, compared to 43% for Dual-Agent LLM and 42% for XGBoost. The advantage is most pronounced at small $k$ values, where missed detections are particularly costly.

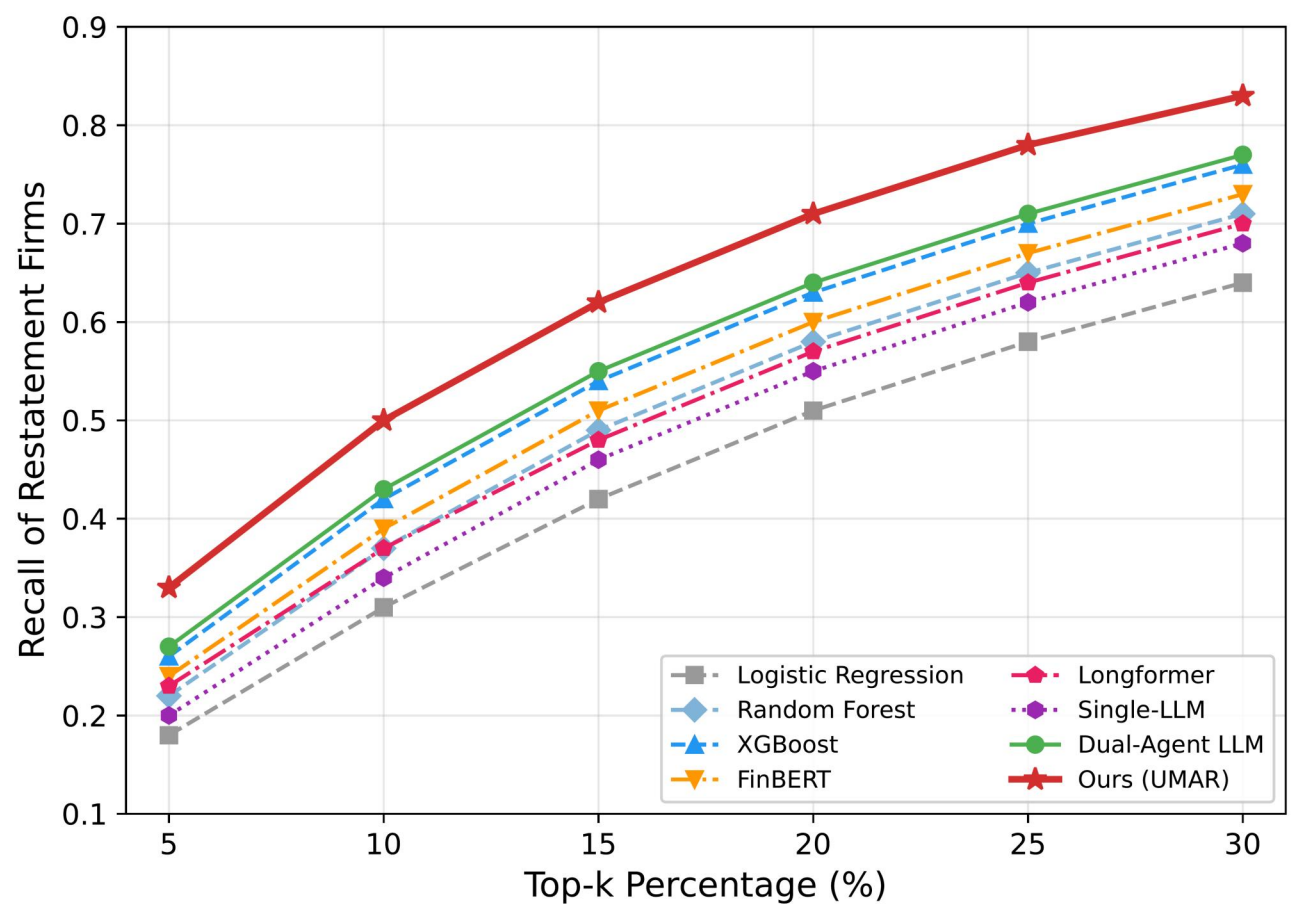

Fig. 2. Top- $k$ recall of restatement firms across methods. UMAR consistently identifies a higher proportion of restatement firms at every inspection threshold.

## E. Evidence Conflict and Calibration Analysis

Fig. 3(a) examines restatement rates across evidence agreement patterns using cross-validated out-of-fold predictions over all 3,200 observations. When all agents agree on high risk, the restatement rate is 52% ($n = 52$); when Text and Financial agents agree while CAM indicates low risk, the rate is 33% ($n = 85$). When all agents agree on low risk, the rate drops to 5.0% ($n = 2,783$), well below the 8.2% base rate, confirming effective low-risk screening. The "CAM High, Text-Fin Low" pattern (25%, $n = 68$) highlights cases where auditor-disclosed risks are not yet reflected in other evidence, underscoring CAM's complementary value.

Fig. 3(b) shows that UMAR's predicted probabilities align more closely with observed restatement frequencies than baselines, which exhibit systematic underconfidence in middle probability ranges. This improvement stems from Dempster–Shafer fusion explicitly modeling residual uncertainty.

## F. Ablation Study

Table III presents ablation results. Removing the Financial Ratio Agent causes the largest AUROC drop ($-0.038$), followed by the MD&A Text Agent ($-0.029$) and the CAM Agent ($-0.021$). Replacing Dempster–Shafer fusion with simple averaging reduces AUROC by 0.024 and increases ECE from 0.052 to 0.081, confirming the importance of principled uncertainty fusion.

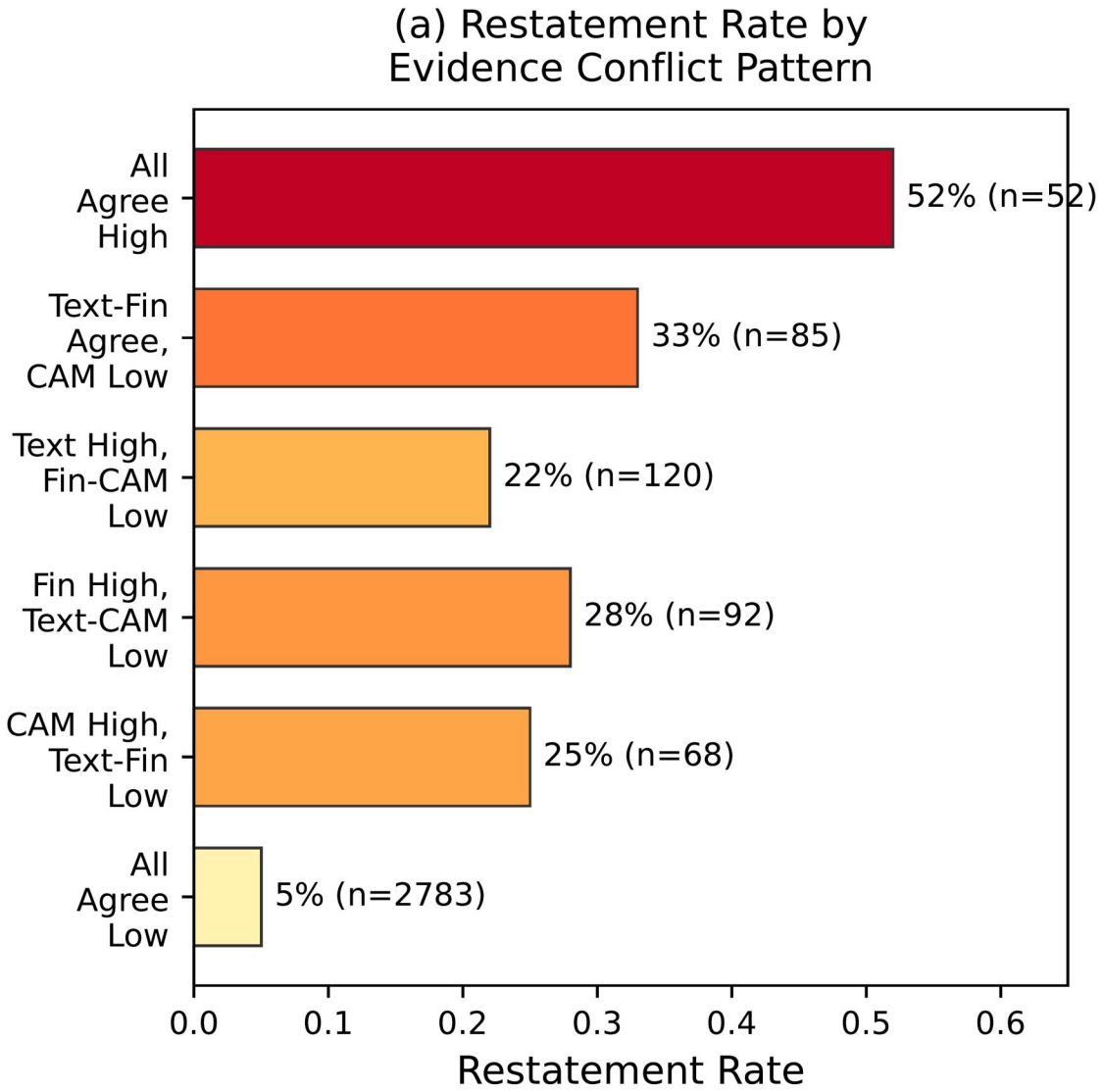

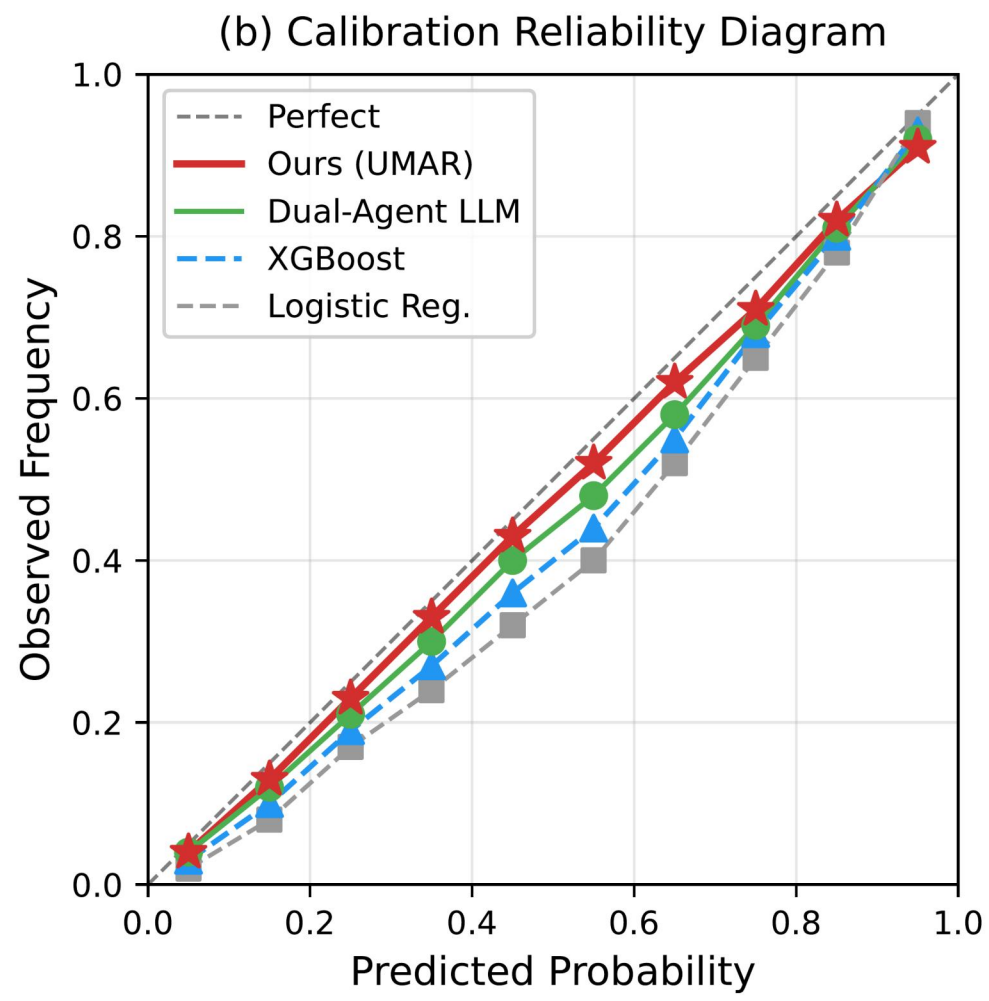

Fig. 3. (a) Restatement rate by evidence conflict pattern based on cross-validated out-of-fold predictions over all 3,200 observations. All-agree-high cases show 52% restatement rate versus 5.0% for all-agree-low (overall base rate: 8.2%). (b) Calibration reliability diagram showing UMAR achieves improved calibration (ECE = 0.052) compared to baselines.

TABLE III.
ABLATION STUDY RESULTS ON THE 2023 TEST SET.

| Variant | AUROC | PR-AUC | Brier↓ | ECE↓ |
|---|---|---|---|---|
| Full UMAR | 0.782 | 0.341 | 0.068 | 0.052 |
| w/o Text Agent | 0.753 | 0.312 | 0.073 | 0.063 |
| w/o Financial Agent | 0.744 | 0.298 | 0.075 | 0.068 |
| w/o CAM Agent | 0.761 | 0.325 | 0.071 | 0.059 |
| Avg. fusion (no D–S) | 0.758 | 0.318 | 0.074 | 0.081 |

## V. CONCLUSION

We presented UMAR, an uncertainty-aware multi-agent framework for audit risk assessment that integrates MD&A text, financial ratios, and critical audit matters through Dempster–Shafer evidence fusion. Our experiments on U.S. SEC filings demonstrate consistent improvements over existing baselines, with notable gains in calibration and evidence conflict detection. Limitations include restriction to a single dataset and the lack of validation in a live audit setting. Future work should extend the evaluation to other markets and explore integration within interactive audit planning systems.